# MaxGlaViT: A novel lightweight vision transformer-based approach for early diagnosis of glaucoma stages from fundus images

Mustafa Yurdakul[1], Kübra Uyar[2]*, Şakir Taşdemir[3]

[1] Kırıkkale University, Computer Engineering Department, Kırıkkale, Turkey
[2] Alanya Alaaddin Keykubat University, Computer Engineering Department, Antalya, Turkey
[3] Selcuk University, Computer Engineering Department, Konya, Turkey

**Abstract**

**Background and objective:** Glaucoma is a prevalent eye disease that progresses silently without symptoms. If it is not detected and treated at early stages, it can lead to permanent vision loss. Computer-assisted diagnosis (CAD) systems are crucial in timely and efficient eye disease identification. Deep learning-based CAD systems have become valuable tools in early detection and treatment.

**Methods:** In this study, a novel lightweight model based on the restructured Multi-Axis Vision Transformer (MaxViT), MaxGlaViT, was designed for early detection of glaucoma stages. Firstly, MaxViT was scaled to optimize the number of blocks and channels of the model, resulting in a lighter architecture. Secondly, the stem in the MaxViT was improved by adding various attention mechanisms (CBAM, ECA, SE) after the convolution layers. As a result, a model was obtained that learns complex features more efficiently. In the third stage, the MBConv structures in the MaxViT blocks were replaced by advanced DL blocks (ConvNeXt, ConvNeXtV2, InceptionNeXt). The model was evaluated using the Harvard Dataverse V1 (HDV1) dataset that contains fundus images belonging to different glaucoma stages. In the experimental studies, state-of-the-art 40 Convolutional Neural Networks (CNN) and 40 ViT models were also evaluated on the HDV1 dataset to prove the efficiency of the proposed MaxGlaViT model.

**Results:** Among CNN models, EfficientB6 outperformed all other CNN models with an accuracy of 84.91%. On the other hand, among ViT models, MaxViT-Tiny performed the best with an accuracy of 86.42%. Then, the scaled MaxViT achieved an accuracy of 87.93%. With the addition of ECA to the stem block, the accuracy increased to 89.01%. Another improvement was achieved by replacing the MBConv structure in the MaxViT block with ConvNeXtV2, with an accuracy of 89.87%. In line with all these results, scaled MaxViT was reconstructed using ECA in the stem block and ConvNeXtV2 in MaxViT block achieved an accuracy of 92.03%.

**Conclusions:** By testing 80 DL models for diagnosing glaucoma stages from fundus images, the proposed study is further expanded as the most comprehensive and comparative attempt in the current literature. In comparison with experimental and state-of-the-art methods, MaxGlaViT demonstrates notable performance, achieving 92.03% accuracy, 92.33% precision, 92.03% recall, 92.13% f1-score, and 87.12% Cohen's kappa score.

*Corresponding Author: Kübra Uyar (kubra.uyar@alanya.edu.tr), Alanya Alaaddin Keykubat University, Computer Engineering Department, 07425, Antalya, Turkey.

## 1. Introduction

The eye is a complex and unique organ that enables humans and many other forms of living beings to see. It helps us to understand the world around us by detecting light. The eye forms images by refracting and focusing light, a process that takes place in retina located at the back of the eye. Thanks to light cells (rod and cone cells), the retina converts images into electrical signals. Then these signals are transmitted via the optic nerve to the brain, where they are interpreted and transformed into an image. Eye health is crucial for keeping this process flowing smoothly, and any problems can affect our ability to see. Glaucoma is a chronic eye disease that occurs due to increased intraocular pressure and causes blindness by damaging the optic nerve head. In the early stages of the disease, patients do not exhibit symptoms of vision loss, while in its advanced stages, vision loss becomes more apparent. Glaucoma known as silent theft of eyesight is an incurable disease; however, with early diagnosis and medication, its progression can be prevented. Ophthalmologists require a detailed digital image of the eye to diagnose glaucoma. Therefore, the structural changes in the optic disc, nerve loss and atrophy in the peripapillary region are examined from the fundus image which is a medical imaging technique that displays the structure of the eye in color [1-3]. However, the number of ophthalmologists worldwide is insufficient, and the existing specialists are working under heavy workloads. Moreover, the correct interpretation of the details and making the correct diagnosis requires experience. For all these reasons, there is a need for a CAD system that leverages advanced algorithms to analyze complex medical imaging data and supports specialists. There are numerous studies and approaches aimed to diagnose glaucoma by analyzing fundus images. One of the CAD methods is the classification of medical images based on feature extraction and machine learning (ML) algorithms. Nayak et al. [4] extracted the cup-to-disc ratio, the ratio of the distance between the center of the optic disc and the optic nerve head to the diameter of the optic disc, and the ratio of the area of blood vessels from fundus images labeled as normal and glaucoma. They classified the features using an artificial neural network (ANN) and achieved a recall and specificity of 100% and 80%, respectively. Bock et al. [5] extracted a number of key features from fundus images using raw intensities, Fourier analysis, and spline interpolation to determine the Glaucoma Risk Index (GRI). Principal Component Analysis (PCA) was used to dimensionally reduce the features and classify them using support vector machines (SVM) with an accuracy of 80%, recall of 73%, and a specificity of 85%. Acharya et al. [6] used a combination of texture and higher order spectral (HOS) features of the fundus image and the Random Forest (RF) algorithm to determine the GRI and achieved a classification accuracy of 91.7%. Acharya et al. [7] used Gabor transformation and PCA approaches for feature extraction to classify normal and glaucoma images. Naive Bayes (NB) and SVM were tested and the most successful result was achieved with SVM with an accuracy of 93.10%, recall of 89.75%, and specificity of 96.20%. In particular, studies mentioned in the literature review above (4-7) work on glaucoma disease datasets as the assessment indicator. Table 1 summarizes the feature extraction and ML classification studies for glaucoma diagnosis.

**Table 1.** Summary of feature extraction and ML classification studies in the literature on glaucoma diagnosis.

| Reference study | Features and techniques | Classifier | Performance metrics |
|---|---|---|---|
| Nayak et al. [4] | Cup-to-disc ratio<br>Ratio of distance between optic disc center and optic nerve head<br>Blood vessel area ratio | ANN | Recall: 100%<br>Specificity: 80% |
| Bock et al. [5] | Raw intensities<br>Fourier analysis<br>Spline interpolation | SVM | Accuracy: 80%<br>Recall: 73%<br>Specificity: 85% |
| Acharya et al. [6] | Texture features<br>HOS features | RF | Accuracy: 91.7% |
| Acharya et al. [7] | Gabor transformation<br>PCA | SVM and NB | Best performance with SVM:<br>Accuracy: 93.10%<br>Recall: 89.75%<br>Specificity: 96.20% |

Feature extraction-based classification techniques require expert knowledge to correctly extract the features from the image, however, this process can be both laborious and error-prone. Manual extraction of features requires engineering skills and domain expertise. Moreover, the features may oversimplify the problem and be temporary, as even experts may miss some important hidden patterns. In addition, the performance of ML algorithms has been insufficient in the face of increasing data [8]. For these reasons, DL algorithms that can automatically extract important features have been proposed. As a deep approach, there has been a wide range of literature studies to analyze and diagnose glaucoma-related eye diseases.

Ahn et al. [9] conducted a comparative study comparing the classification performance of Logistic Regression (LR) and InceptionV3 for multi-stage glaucoma classification, collecting advanced-stage, early-stage, and normal fundus images, referred to as the HDV1 dataset. Using flattened raw pixel features, LR achieved 82.9% training accuracy, 79.9% validation accuracy, and 77.2% test accuracy. InceptionV3 achieved 99.7% accuracy, 87.7% accuracy on validation data, and 84.5% accuracy on test data. Juneja et al. [10] proposed G-Net, a modified version of U-Net, to classify fundus images as glaucomatous or non-glaucomatous. The G-Net architecture was basically inspired by U-Net, scaling U-Net and optimizing the hyperparameters. They obtained 95.8% accuracy in an experimental study on the DRISHTI-GS dataset consisting of 101 images. Similarly, in another study by Juneja et al. [11], a CNN model called CoG-NET, a modified version of Xception, was proposed to classify fundus images as normal and abnormal, and 93.5% accuracy, 95% recall, and 99% specificity were achieved. Chai et al. [12] proposed a multi-branch approach for glaucoma diagnosis from fundus images using a combination of CNN, regional convolutional neural network (R-CNN), and fully convolutional network (FCN). In the first branch, the image was taken as input and features were extracted through a CNN. In the second branch, a faster R-CNN was used to obtain the optical disc region. In the third branch, an FCN model was used to segment the disc area, dish area, and peripapillary atrophy area and then compute the measurements. As a result, their approach achieved 91.51% accuracy, 92.33% recall, and 90.90% specificity on a custom dataset. Haouli et al. [13] compared the classification performances of ViT (B16, B32, L16, L32) and CNN (Xception and ResNet152V2) models for binary (normal-abnormal) glaucoma classification, inspecting the batch size effect on performance. They combined five datasets which are ACRIMA, RIM-ONE, Drishti-GS1, HRF, and SJCHOI86-HRF, achieving the best result with an accuracy of 92.67% using ViT-L32 on the combined dataset. Das et al. [14] proposed a model named adapter and enhanced self-attention network (AES-Net) to improve the performance of CNN models for glaucoma diagnosis. Firstly, the authors tested various CNN models on HDV1 and LMG datasets. The DenseNet169 was the most successful model on the HDV1, achieving 83.83% accuracy, 83.35% precision, 83.87% sensitivity, and 83.41% f1-score. Similarly, it achieved 82.80% accuracy, 82.58% precision, 82.80% recall, and 82.61% f1-score on the LMG. The DenseNet169 was used as the backbone and the self-attention mechanism enhanced with the proposed adapter was added to the last feature layer of DenseNet169. With these modifications, an accuracy of 86.20%, precision of 85.32%, recall of 85.77%, and f1-score of 85.46% were achieved on the HDV1. On the LMG dataset, 84.48% accuracy, 84.27% precision, 84.48% recall, and 84.34% f1-score were achieved. Das and Nayak [15] proposed FJA-Net for multi-stage fundus image classification and tested various CNN models using transfer learning to determine the best backbone. Then, they added a fuzzy joint attention module (FJAM) at the last layer of the best-performing model and compared the classification performance. In experiments on HDV1 and LMG datasets, DenseNet169 achieved 83.83% accuracy, 83.35% precision, 83.83% recall, and 83.41% f1-score on HDV1 and 82.80% accuracy, 82.58% precision, 82.80% recall, and 82.58% f1-score on LMG. When DenseNet169 was used as the backbone and FJAM was added, the accuracy of 87.06%, precision of 87.01%, recall of 87.06%, and f1-score of 86.90% were achieved on the HDV1. On the LMG dataset, 84.91% accuracy, 84.35% precision, 84.91% recall, and 84.55% f1-score were achieved. Das et al. [16] proposed a novel cascaded attention-based network model called CA-Net for efficient multi-stage glaucoma classification. The authors found that DenseNet121 was the most successful model with 83.18% accuracy, 83.10% precision, 83.18% recall, and 83.01% f1-score on the HDV1 and 81.55% accuracy, 82.10% precision, 81.55% recall, and 81.58% f1-score on LMG. Using DenseNet121 with the proposed cascaded attention, the authors obtained 85.34% accuracy, 85.15% precision, 85.34% recall, and 84.92% f1-score on HDV1; 83.85% accuracy, 83.69% precision, 83.85% recall, and 83.48% f1-score on LMG. Das et al. [17] introduced the GS-Net which DenseNet121 model enhanced with a global self-attention module (GSAM) consisting of two parallel attention modules, a channel attention module (CAM) and a spatial attention module (SAM). The authors tested the GS-Net on HDV1 and achieved an accuracy of 84.91% and f1-score of 84.55%.

Table 2 lists the detailed summary of the key studies within the categories of DL-based and ViT-based methods in the literature related to glaucoma diagnosis. This table also helps to illustrate the progress and effectiveness of various methodologies in the field of glaucoma stage detection.

The analysis of previous studies highlights the significant advancements made in both ML and DL approaches for glaucoma detection. Tables 1 and 2 show that there is a continuous trend in the literature to improve the performance of models for glaucoma-level detection. Despite these advances, challenges remain in generalizability, interpretability, and scalability of existing solutions. Overcoming these challenges requires the development of new approaches that not only improve classification performance but also preserve the efficiency of the model for practical applications. In our work, we build on these findings and present innovative improvements to provide higher diagnostic accuracy and reduce model complexity.

**Table 2.** Summary of studies in the literature related to diagnosing glaucoma using DL algorithms.

| Study | Model | Method | Performance metrics |
| --- | --- | --- | --- |
| Ahn et al. [9] | LR and InceptionV3 | Pixel-based features and transfer learning-based classification | LR: Accuracy: 77.2% InceptionV3: Accuracy: 84.5% |
| Juneja et al. [10] | G-Net | Scaled U-Net, optimized hyperparameters | Accuracy: 95.8% |
| Juneja et al. [11] | CoG-NET | Modified Xception architecture | Accuracy: 93.5% Recall: 95% Specificity: 99% |
| Chai et al. [12] | Multi-Branch Approach (CNN, Faster R-CNN, FCN) | Multi-branch architecture combining different models | Accuracy: 91.51% Recall: 92.33% Specificity: 90.90% |
| Haouli et al. [13] | ViT and CNN models | Comparative study | Best result with ViT-L32 Accuracy: 92.67% |
| Das et al. [14] | AES-Net | Added adapter and enhanced self-attention to DenseNet169 | Accuracy: 86.20% (HDV1) 84.48% (LMG) |
| Das and Nayak [15] | FJA-Net | FJAM added to DenseNet169 backbone | Accuracy: 87.06% (HDV1) |
| Das et al. [16] | CA-Net | DenseNet121 with the cascaded attention | Accuracy: 85.34 % (HDV1) 83.85% (LMG) |
| Das et al. [17] | GS-Net | Global self-attention module added to DenseNet121 | Accuracy: 84.91% (HDV1) |

The following describe the contributions and novelty of our study:
- A comprehensive literature review on artificial intelligence techniques for glaucoma stage detection is presented.
- This paper reports a most extensive comparison covering 40 CNN and 40 ViT models to detect glaucoma stages.
- The channel and block numbers of MaxViT are rescaled, resulting in a lightweight architecture that provides high and robust accurate results.
- The stem block of MaxViT is improved with an attention mechanism with efficient channel attention (ECA) after a comparison with the convolutional block attention module (CBAM) and squeeze-and-excitation (SE).
- The MBConv block in MaxViT is replaced by the ConvNeXtV2 comparing ConvNeXt and InceptionNeXt modules. This change results in a model with improved generalization capabilities on test data.

- The proposed MaxViT-based model named MaxGlaViT enhances glaucoma stage detection by improving the stem and MaxViT block with attention modules and advanced convolutional blocks.
- In addition to a comparative performance analysis between MaxGlaViT series and various state-of-the-art CNN and ViT models, MaxGlaViT's performance is compared with other literature studies.
- Experimental results demonstrate that MaxGlaViT outperforms recent models in literature and surpasses a total of 80 DL models.

The rest of this paper is organized as follows: Section 2 explains the material and method of the study. The proposed framework and details are mentioned in Section 3. The experimental results and discussion are detailed in Section 4. In addition to this, the comparison of the proposed approach with other alternative approaches is reported in this section. Finally, Section 5 summarizes the main findings of the proposed study and gives some future directions.

## 2. Material and Methods

The basic concepts of CNN and ViT, attention modules, advanced DL blocks, the dataset description procedure, and various performance measurement metrics are included in the following subsections.

### 2.1. CNN

CNN is a DL algorithm widely used in image analysis studies such as image classification, detection, and segmentation. CNNs are designed to capture spatial hierarchies in data, using convolutional layers to extract local patterns and pooling layers to reduce dimensionality, allowing the network to learn features increasingly through each layer. CNNs are structured with layers that apply filters to input images, progressively detecting more complex features through deeper layers.

In the literature, there are various CNN models that contain different topologies such as dense block, ghost module, inception block, residual block, and separable convolution. Various CNN architectures have been employed utilizing transfer learning to improve model performance for glaucoma stage detection. DenseNet models that contain dense blocks; EfficientNet models that include inverted residual blocks; GhostNet and variants that use ghost modules; Inception models with various versions that include inception modules; ResNet, MobileNet, and NASNetMobile models that contain residual blocks; VGG models that contain stacked blocks (basic CNN layers); and Xception model that contains separable convolution were all assessed as CNN models.

### 2.2. ViT

Transformer is a recent DL algorithm that derives its power from self-attention and was first used in natural language processing (NLP). Vaswani et al. [18] used transformers for machine translation, Devlin et al. [19] proposed the BERT model, a bidirectional language representation that takes context into account. Brown et al. [20] proposed the generative pre-trained transformer

GPT, a large language model. And so, transformer-based models kick-started a new era in the field of NLP. Following these studies, researchers started to apply the transformer models to image analysis. The first ViT model proposed by Dosovitskiy et al. [21] achieved 88.36% accuracy on ImageNet by directly applying transformers to image patches. Subsequently, many ViT models have been developed to carry out various image-based tasks.

Pyramid vision transformers (PVT) [22] process images in a pyramid structure to produce multi-feature maps at different resolution levels to detect both small and large-scale objects. Swin transformer [23] effectively combines local and global context with the shifted window technique. The image is divided into fixed windows and information is shared through shifts between the windows. Dual-axis vision transformer (DaViT) [24] analyzes images in both horizontal and vertical axes for more comprehensive feature extraction. FastViT [25] is a speed-oriented ViT model designed for efficient computer vision tasks, balancing both accuracy and speed. It combines the strengths of transformers and convolutions, leveraging sparse attention and advanced compression techniques to reduce computational load. Global context vision transformer (GCViT) [26] is a model developed to capture global context information. It is particularly effective for high-resolution vision tasks like image classification, object detection, and segmentation, optimizing computational resources without sacrificing model accuracy. FlexiViT [27] is a model that randomizes the patch sizes in ViT models, providing high performance for different patch sizes with a single set of weights. The method offers flexibility and computational efficiency by eliminating the need to train separate models for different patch sizes. GPViT [28] is a non-hierarchical ViT model that can efficiently perform global knowledge transfer over high-resolution features. GPViT uses an innovative group propagation block that allows information transfer by grouping features and then returns the information back to the initial features. LeViT [29] is a different model that is fast and efficient by combining CNN principles with transformer architecture. It uses hybrid architecture elements, combining convolutions with transformers to reduce computational complexity and improve processing speed. Finally, MaxViT [30] is another ViT that integrates local and global context using global and block attention mechanisms. It is a model that integrates transformers with convolutions and advanced attention mechanisms to achieve both high efficiency and accuracy in visual tasks.

### 2.3. Attention Modules

In recent years, the attention mechanism has gained popularity and has been commonly used to improve the accuracy of DL models [31,32]. The attention mechanism, in simple terms, detects which areas in the feature map are more important and so the model focuses on these areas. Injection of attention into convolution blocks is one of the implementation techniques, showing great potential for performance improvement in many studies [33,34]. Within the scope of this study, the popular attention mechanisms (CBAM, ECA, and SE) were used in the stem block of the MaxViT model to improve performance. The general attention mechanisms are explained in the following subsections.

*2.3.1. SE*

SENet which uses the SE module is a DL model that aims to improve the performance of models by using a customized attention mechanism to determine the importance of channels in feature maps [35]. SE module, transforms the input feature map, compressing its dimensions into a compressed format while maintaining the number of channels. In the process of compressing the feature map, global average pooling (GAP) is applied to obtain a vector representing each channel. The vector is of size 1x1xC and reflects the intensity of each channel. Then, the vector is processed with fully connected layers and activation functions to generate attention scores that represent the importance of the channels. As a result of that process, scaling coefficients of size 1x1xC are obtained. In the final stage, these coefficients are applied to the initial feature map on a channel-by-channel basis. Each channel is rescaled according to its attention coefficient so that the model makes the important channels more salient and the unimportant ones weaker. The schematic diagram of the SE module is shown in Fig. 1.

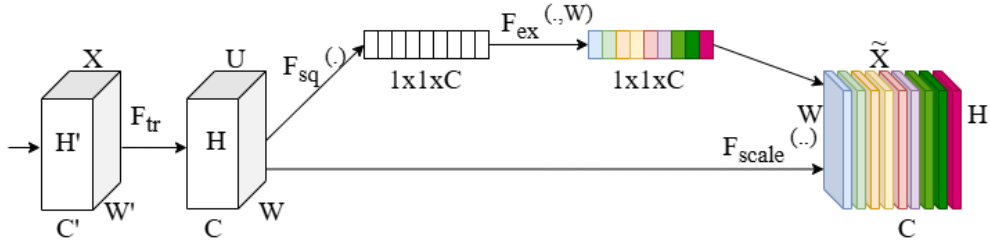

**Fig. 1.** SE module.

*2.3.2. ECA*

Inspired and in pursuit of improving SENet, Wang et al. found that dimensionality reduction has a side effect on channel attention based on empirical studies [36]. They proposed ECA, which avoids dimensionality reduction and captures cross-channel interaction in an efficient way. The ECA first applies the GAP to the input tensor. With GAP, average values are calculated for each channel to reduce the spatial dimensionality and the tensor is transformed into $1x1xC$. Then, ECA uses one-dimensional convolution to learn the channel relationships. The resulting channel attention map is multiplied by the input feature map on a channel-by-channel basis to scale the importance of each channel so that the model emphasizes the important channels. The schematic diagram of the ECA module is shown in Fig. 2.

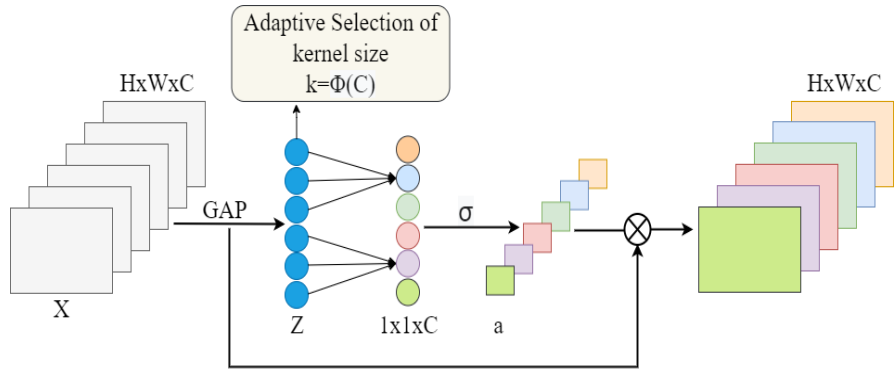

**Fig. 2.** ECA module.

*2.3.3. CBAM*

CBAM [37] generates attention maps in two stages by analyzing the feature maps. In the first stage, the channel attention module compresses the spatial dimension of the input feature map to determine how important each channel is. An average and maximum pooling method is used to create two different context descriptors. Then, the descriptors are processed through a shared neural network, resulting in a channel attention map.

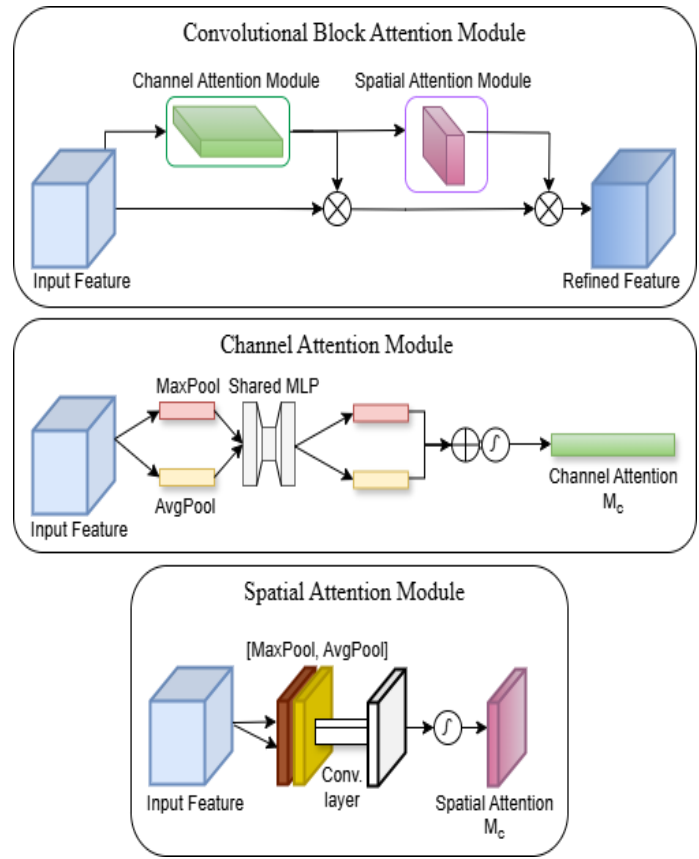

**Fig. 3.** CBAM module.

In the second stage, the spatial attention module collects cross-channel information to understand which regions are more important. Then, two two-dimensional maps are generated by averaging and maximum pooling. These maps are combined to produce a spatial attention map through a convolution layer. This process improves the quality of the final output by determining which features the model should focus on. The schematic diagram of the CBAM is shown in Fig. 3.

**2.4. Advanced DL Blocks**

Advanced DL blocks are specialized architectural components in DL models designed to enhance learning efficiency, scalability, and model performance. Advanced blocks refine feature extraction while reducing the computational cost. These blocks are modular and can be stacked or combined with other architectures, making them versatile tools for creating novel powerful models. The general advanced DL blocks are explained in the following subsections.

*2.4.1. ConvNeXt and ConvNeXtV2*

ConvNeXt [38] is a modern CNN architecture that modernizes classical convolutional designs to achieve competitive performance on image classification tasks, competing with ViTs. Developed with insights from both CNNs and ViTs, ConvNeXt refines traditional convolutional layers to improve efficiency, scalability, and accuracy. Simplified convolutional blocks, large kernels for the expanded receptive field, layer normalization and gaussian error linear unit (GELU) activation, inverted bottleneck design, and hierarchical feature representation are a breakdown of the key features and innovations in ConvNeXt. It uses large kernel sizes of 7x7 to scan a larger area and provide efficient information capture. It uses up-to-date techniques such as layer normalization and GELU activation function. It offers a simple yet effective structure with depthwise convolution for downsampling followed by 1x1 convolution layers. In the ConNeXtV2 [39] model, which is an evolution of ConvNeXt, the LayerScale layer is removed from the block and the GRN module is added to handle feature changes and avoid feature collapse in the learning process. The GRN layer increases the contrast between channels, effectively improving the performance of the model.

*2.4.2. InceptionNeXt*

Yu et al. [40] proposed InceptionNeXt with the considering baseline as ConvNeXt. Compared to ConvNeXt, InceptionNeXt is an architecture used to make large-kernel convolutions more efficient. In the InceptionNeXt architecture, the feature map is provided as input to an inception block so that comprehensive feature extraction can be performed with filters of different kernel sizes. The extracted features are then combined and normalized. Finally, the feature map provided as input is merged with the final feature map with the residual block to obtain the resultant features. Fig. 4 illustrates the general structures of the ConvNeXt, ConvNeXtV2, and InceptionNeXt.

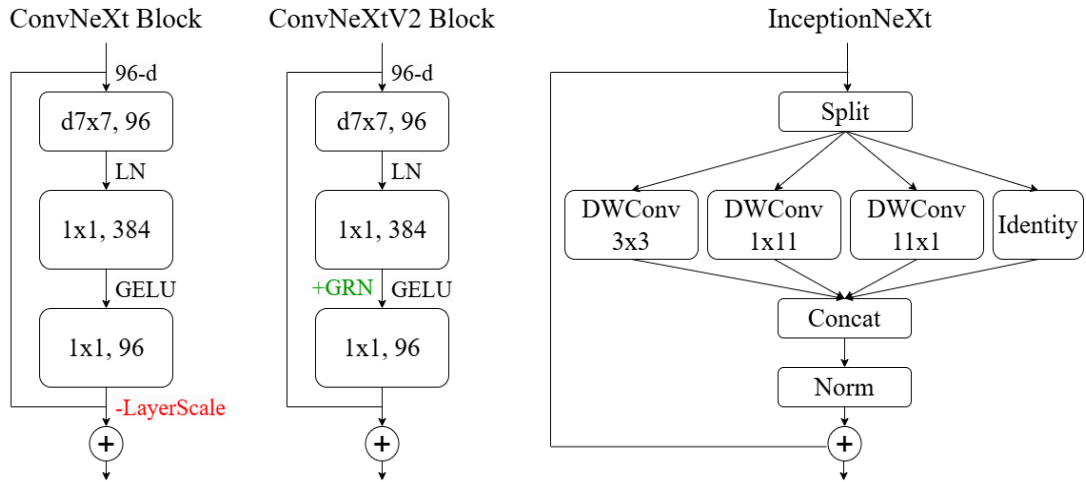

**Fig. 4.** Advanced DL blocks.

## 2.5. MaxViT

MaxViT is a state-of-the-art DL architecture that effectively merges the strength of CNN and ViT. Introduced by Tu et al. [30], MaxViT employs a hybrid approach that integrates both convolutional layers and self-attention mechanisms. This design allows the model to capture local features through convolutional operations while also leveraging the global context provided by attention layers. The MaxViT model achieved 86.5% top-1 accuracy on ImageNet-1K and 88.7% top-1 accuracy on ImageNet-21K.

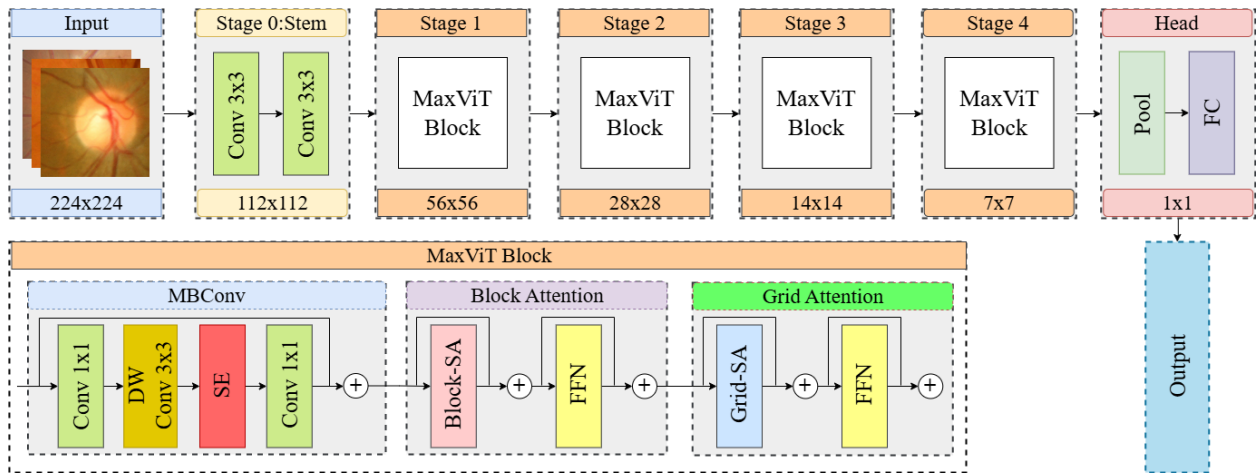

**Fig. 5.** An illustration of the MaxViT architecture.

MaxViT consists of three main modules: stem block, MaxViT block, and classifier head as shown in Fig. 5. The input image is transmitted to the stem block for feature extraction. In the stem block, there are two convolution layers that extract features from the image. The first convolution layer uses a 3x3 filter and is implemented with a stride of 2 units. Then, a second convolution layer with another 3x3 filter, using a stride of 1 unit, is applied. It improves the feature extraction by further detailing the feature maps extracted by the first layer. After the convolution layers, a batch normalization (BN) layer is used to stabilize the learning process and GELU activation function is applied to provide a nonlinear structure to the model. In the rest of the model, there are 4 MaxViT blocks and each MaxViT block contains a mobile inverted residual bottleneck convolution (MBConv), a block attention, and a grid attention module.

MBConv first expands the feature map using a 1x1 kernel, which allows the model to learn features over more channels. After the expansion, a 3x3 deep convolution (DW) is applied to the feature map. The DW works on each input channel separately, capturing spatial features along the input. The SE mechanism that calculates the importance of the interdependencies between different feature channels is then applied. In this way, important features are highlighted on the feature map. The final 1x1 convolution in the MBConv structure is used to return the expanded channels back to their original input size. Thus, the basic features learned in the channels expanded with lower parameters are preserved. The feature maps obtained with the MBConv block are given as input to block attention. The block attention module splits the feature map into small windows; a feature map of dimensions HxWxC where H is the height, W is the width and C represents the number of channels, is transformed into a tensor. The feature map is thus divided into C small windows of PxP dimensions. Each window represents a region in the feature map that does not intersect with one another. The self-attention mechanism is then used to understand how the windows are related to each other. The features obtained after the self-attention mechanism are given as input to the feedforward network (FFN). The FFN provides the model to learn more complex patterns and relationships by applying nonlinear transformations to the features. In the last stage of the MaxViT block the grid attention is used. The grid attention module focuses on pixels using a grid evenly distributed over the entire feature map and the grid attention mechanism transforms the feature map. Thereby, the feature map is divided into C times GxG partitions. Self-attention is applied on these partitions and the extracted features are transferred to the FFN as in block attention to learn more complex patterns and relationships by applying nonlinear transformations to the features.

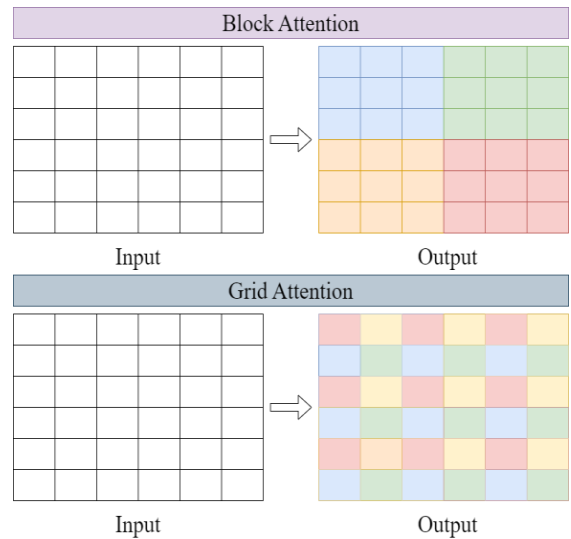

**Fig. 6.** Block and grid attention.

The effects of block and grid attention on feature maps are illustrated in Fig. 6. These attention blocks use residual connections to learn the combination of the original input and the output of the self-attention and feed-forward processes. In this way, the model can learn better and avoid the problem of gradient loss during training. In the last part of MaxViT, the head part, the GAP is applied to the feature map and the resulting feature vector is classified with a fully connected layer.

### 2.6. Description of the Dataset

All the DL models mentioned in this paper have been evaluated using the HDV1 dataset which is proposed in the study [7]. The dataset comprises three distinct classes: advanced glaucoma, early glaucoma, and normal (healthy). Early glaucoma class accounts for 289 samples, followed closely by advanced glaucoma with 467 samples. The normal class included 786 images. All images are 224x224 in size and colorful fundus images. The classes of the data were labeled by means of a consensus decision made by two experts and the dataset was divided into training, validation, and test. The distribution of the data is shown in Table 3. Considering the class imbalance, online data augmentation techniques were applied during training. Fig. 7 presents randomly selected images from the classes in the utilized dataset.

**Table 3.** The number of images in the train, validation, and test sets of HDV1.

| Class name (glaucoma stage) | Train | Validation | Test | Total |
| --- | --- | --- | --- | --- |
| Advanced | 228 | 98 | 141 | 467 |
| Early | 141 | 61 | 87 | 289 |
| Normal | 385 | 165 | 236 | 786 |
| Total | 754 | 324 | 464 | 1542 |

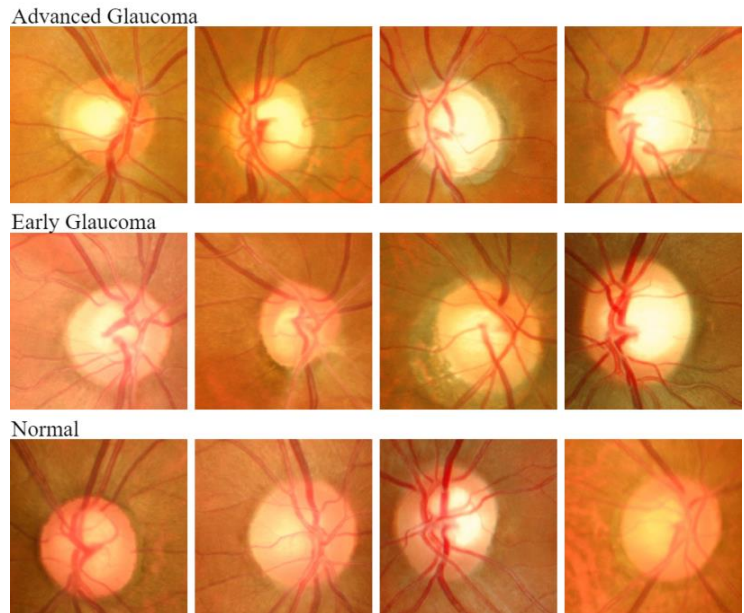

**Fig. 7.** Some fundus images from the HDV1 dataset, showing (top to bottom) advanced, early, and normal classes.

## 2.7. Performance Measurement Metrics

Evaluating the classification success of DL models objectively is an important task. The confusion matrix provides a tabulation of the predictions of model compared to the actual labels. It gives an overview of the performance of the model by distinguishing between correct and incorrect predictions for each class. The table contains values for true positive (TP: samples correctly classified as positive), true negative (TN: samples correctly classified as negative), false positive (FP: samples incorrectly classified as positive) and false negative (FN: samples incorrectly classified as negative).

Accuracy indicates the overall success of the model, precision indicates the effect of false positives, and recall indicates the effect of false negatives. The f1-score is the harmonic mean of precision and recall and better evaluates performance in imbalanced datasets. Cohen's kappa measures the classification success of the model by taking into account the chance factor. With these metrics, it is possible to evaluate the model from different perspectives and to better understand its strengths and weaknesses. The mathematical formula of the performance metrics is given in Eq. 1-5.

$$Accuracy = \frac{TP + TN}{TP + TN + FP + FN} \tag{1}$$

$$Precision = \frac{TP}{TP + FP} \tag{2}$$

$$Recall = \frac{TP}{TP + FN} \tag{3}$$

$$F1 - score = \frac{2 * Precision * Recall}{Precision + Recall} \tag{4}$$

$$Cohen's\ kappa = \frac{p_0 - p_e}{1 - p_e} \tag{5}$$

In Eq. 5, $p_0$ is defined as the observed proportion of agreement, which represents the relative frequency of cases where the raters agree. $P_e$ is the expected proportion of agreement, calculated on the basis of the frequency of each category, and represents the probability of random agreement between the raters.

## 3. Proposed MaxGlaViT Model

Fig. 8 shows the graphical representation of the proposed MaxGlaViT architecture with detailed layers. The proposed MaxGlaViT comprises three main phases: scaling blocks and channels in MaxViT, improving stem block, and enhancing MaxViT block. The first phase optimizes the number of blocks and channels in the MaxViT architecture, which has a direct impact on the computational complexity of the model and the number of parameters. In this way, a lightweight architecture is obtained without compromising the performance of the model and reducing the computational cost. In the second stage, various attention modules were added after the convolutional layers in the stem block in the MaxViT architecture. Finally, by replacing the MBConv blocks in the MaxViT block with advanced CNN blocks, the proposed model was constructed. The implementation details of the construction phases of the proposed MaxGlaViT explained in the following subsections.

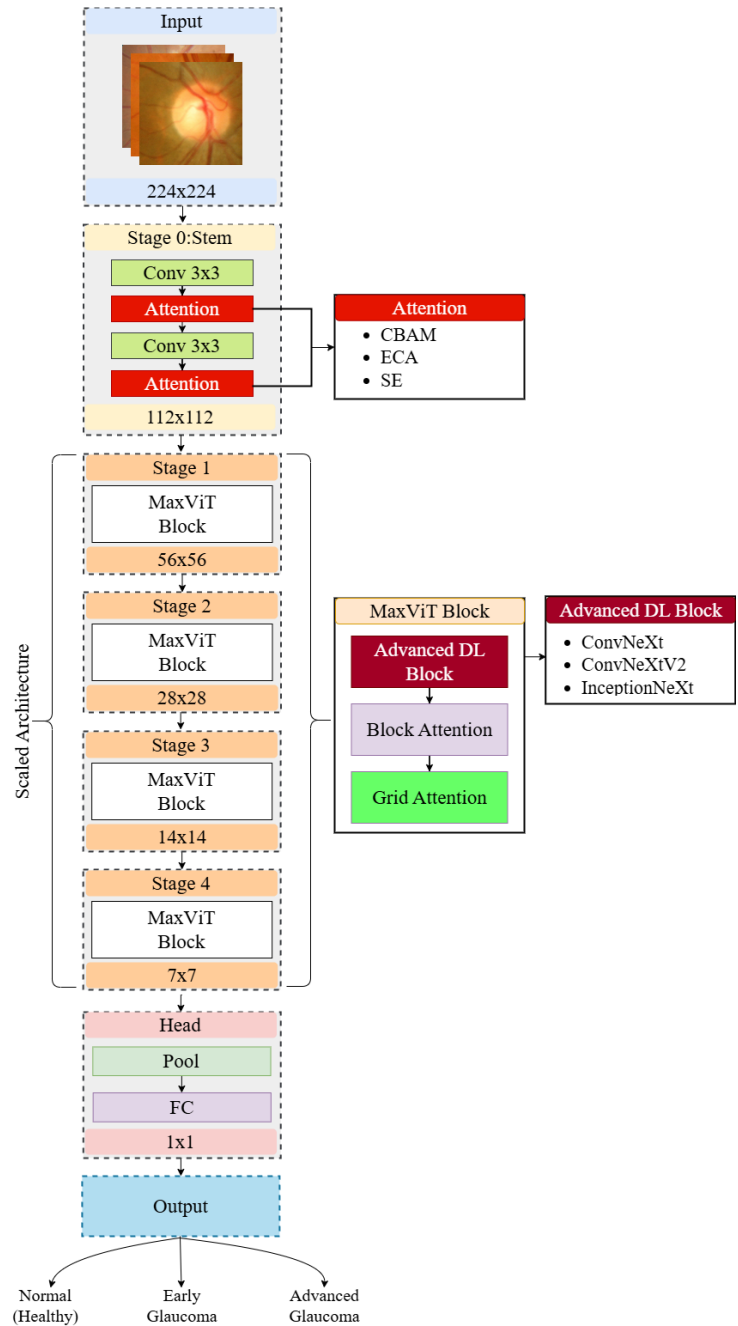

**Fig 8.** The graphical abstract of the designed MaxGlaViT for multiple glaucoma stage detection.

### 3.1. Scaling MaxViT

The scalability of ViTs makes them flexible, enabling them to be used effectively on datasets of different sizes and for a variety of visual tasks. Furthermore, the ability to increase or decrease the number of parameters with scaling helps the model to adapt to different hardware capacities. Recent studies show that the performance of MaxViT can be improved by carefully optimizing the model parameters [41- 43]. In particular, the number of blocks and channels has a direct impact on the classification performance and computational complexity of the model. While the number of blocks increases the depth of features that the model can learn at each level, excessive block usage increases the computational load and memory consumption. However, an excessive number of blocks and channels not only increases hardware burden but can also lead to model overfitting, which can negatively impact classification performance. In addition, if there is an imbalance between data and high-capacity models, there is a tendency for the model to overlearn the data, resulting in a less generalized and poorer-performing model.

For the purposes of this study, the number of blocks and channels of the stem block, while remaining in its original form, were set to block 2 and channel 32 for stage 1, block 2 and channel 64 for stage 2, block 2 and channel 128 for stage 3, and block 2 and channel 256 for stage 4. These values have been chosen as optimal considering the amount and structure of the data. Table 4 lists the block and channel values and the total number of parameters for each stage of the MaxViT Tiny, Base, Small, and Large models. With the scaled model, a model with 6.2M parameters was obtained with 80% fewer parameters than the Tiny version (31M).

**Table 4.** MaxViT architecture variants (B and C denote the number of blocks and number of channels for each stage).

| Stage | Scaled MaxViT | MaxViT-Tiny | MaxViT-Base | MaxViT-Small | MaxViT-Large |
|---|---|---|---|---|---|
| Stem Block (Stage 0) | B = 2, C = 64 | B = 2, C = 64 | B = 2, C = 64 | B = 2, C = 64 | B = 2, C = 128 |
| MaxViT Block (Stage 1) | B = 2, C = 32 | B = 2, C = 64 | B = 2, C = 96 | B = 2, C = 96 | B = 2, C =128 |
| MaxViT Block (Stage 2) | B = 2, C = 64 | B = 2, C =128 | B = 6, C = 192 | B = 2, C = 192 | B = 6, C = 256 |
| MaxViT Block (Stage 3) | B = 2, C = 128 | B = 5, C = 256 | B = 14, C = 384 | B = 5, C = 384 | B = 14, C = 512 |
| MaxViT Block (Stage 4) | B = 2, C = 256 | B = 2, C = 512 | B = 2, C = 768 | B = 2, C = 768 | B = 2, C = 1024 |
| Parameter Count (M: Million) | 6.2 | 31 | 119 | 69 | 212 |

### 3.2. Improving Stem and MaxViT Block

In MaxViT, the stem block is used to extract features from the image for the first time, and the extracted features are processed by MaxViT blocks. Since the features extracted in the stem block are used as input to other blocks, it is crucial to improve the stem block. To increase the feature representation capacity of the stem block, an attention module was added after each convolutional layer. The first attention module was placed after the first convolution layer, enabling more efficient processing of the extracted feature maps, and the second module was implemented after the second convolution layer, enabling the network to learn more complex detailed features. CBAM, ECA, and SE modules were used in the experiments to determine the attention that would best improve the performance of the model. Another improvement is conducted in the MaxViT block. The MBConv blocks in the original structure of the MaxViT block perform local feature extraction, replacing them with more advanced convolution modules significantly improves the performance and generalization capability of the model. Modern convolution modules enable deeper and more meaningful feature extraction, strengthening the model's learning capacity and optimizing its performance. In this study, state-of-the-art ConvNeXt, ConvNeXtV2, and InceptionNeXt modules are experimentally used instead of MBConv blocks, respectively. The goal of this modification is to provide a more efficient and flexible learning process by improving the accuracy, robustness, and overall performance of the model.

## 4. Experimental Results and Discussion

This section provides a comprehensive assessment of the advanced DL models for glaucoma stage detection on test data. The performance metrics and outcomes demonstrated in the confusion matrix associated with the proposed model were discussed.

The experiments were performed on a computer with 128 GB of RAM, two Nvidia RTX 3090 24GB GPUs combined with an NVLink bridge, and an Intel i9 processor. Python was utilized as the programming language and Keras, a sub-library of Tensorflow, was used to perform CNN, ViT, and MaxGlaViT models.

All DL models were trained with a transfer learning method based on ImageNet data weights. Throughout the experiments, a standardized configuration was consistently applied to all models. Each model was trained with categorical cross-entropy loss and the Adam optimizer with a learning rate of 1e-3 and a weight reduction value of 0.8. The batch size was set to 16 and the number of training epochs to 50 in all experiments. Furthermore, data augmentation techniques such as scaling, rotation, and vertical-horizontal shifting were used to reduce overfitting. These hyperparameters were selected in the same way as in recent work [14] for a fair comparison.

### 4.1. Results of the CNN Models

This section discusses the performance metrics of the various CNN models after the test process and the outcomes depicted in the confusion matrix. Table 5 presents the experimental results of various CNN models for glaucoma stage detection on test data.

As listed in Table 5, the accuracy values of CNN models vary between 67.24% and 84.91%. Among the GhostNet series models, GhostNet100 showed the highest performance with 81.43% accuracy and 81.42% f1-score. GhostNet-130 was the most successful model after GhostNet100 with 80.80% accuracy and 80.81% f1-score. In contrast, GhostNetV2100 performed the worst with 78.44% accuracy and 78.11% f1-score. In the DenseNet series, DenseNet201 model achieved the best result with 84.05% accuracy and 83.97% f1-score. DenseNet169 and DenseNet121 were competitive with 82.33% and 83.41% accuracy. The EfficientNet series models achieved high accuracy and f1-score values with different configurations. The EfficientNetB6 performed the best with an accuracy of 84.91% and an f1-score of 85.25%. EfficientNetB5 also performed well with 84.27% accuracy and 84.10% f1- score. EfficientNetB7 showed high performance with 84.82% accuracy and 84.98% f1-score, but it was slightly lower than EfficientNetB6. InceptionResNetV2 and InceptionV3 models also stand out with their high accuracy and f1-score values. InceptionResNetV2 outperforms the other model with 82.54% accuracy and 82.85% f1-score. On the other hand, InceptionV3 performs quite well with 81.68% accuracy and 81.51% f1-score. In the MobileNet series, MobileNetV2 had an accuracy of 82.33%, while the NASNet series models underperformed, remaining around 67% accuracy. In the ResNet series, the ResNet50V2 model achieved the best results with 81.90% accuracy and 80.39% f1-score. The Xception model was the best-performing model with 84.70% accuracy and 84.97% f1-score. In the VGG series, VGG16 achieved the best results with 80.60% accuracy and 80.14% f1-score, while the VGG19 model performed lower with 69.18% accuracy. As a result, it was observed that models such as EfficientNetB6, DenseNet201, and Xception showed high performance, while some models such as NASNet showed low performance.

**Table 5.** Results of CNN models.

| Model | Performance measurement metrics (%) | | | | |
|---|---|---|---|---|---|
| | Accuracy | Precision | Recall | F1-score | Cohen's kappa |
| GhostNetV2100 | 78.44 | 77.93 | 78.44 | 78.11 | 64.79 |
| GhostNetV2130 | 81.25 | 80.20 | 81.25 | 80.32 | 68.89 |
| GhostNetV2160 | 80.17 | 79.03 | 80.17 | 78.87 | 66.85 |
| GhostNet050 | 80.96 | 81.32 | 80.96 | 81.11 | 69.15 |
| GhostNet100 | 81.43 | 81.78 | 81.43 | 81.42 | 69.68 |
| GhostNet130 | 80.80 | 81.05 | 80.80 | 80.81 | 68.61 |
| DenseNet121 | 83.41 | 82.95 | 83.41 | 83.13 | 72.69 |
| DenseNet169 | 82.33 | 81.60 | 82.33 | 81.75 | 71.03 |
| DenseNet201 | 84.05 | 83.91 | 84.05 | 83.97 | 73.89 |
| EfficientNetB0 | 80.82 | 81.78 | 80.82 | 81.19 | 68.88 |
| EfficientNetB1 | 81.47 | 81.32 | 81.47 | 81.37 | 69.59 |
| EfficientNetB2 | 81.03 | 81.05 | 81.03 | 81.01 | 68.92 |
| EfficientNetB3 | 82.33 | 83.69 | 82.33 | 82.85 | 71.65 |
| EfficientNetB4 | 81.47 | 82.09 | 81.47 | 81.71 | 69.87 |

| Model | | | | | |
|---|---|---|---|---|---|
| EfficientNetB5 | 84.27 | 84.00 | 84.27 | 84.10 | 74.12 |
| **EfficientNetB6** | **84.91** | **85.85** | **84.91** | **85.25** | **75.64** |
| EfficientNetB7 | 84.82 | 85.12 | 84.09 | 84.98 | 75.55 |
| EfficientNetV2B0 | 80.39 | 80.04 | 80.39 | 80.15 | 68.07 |
| EfficientNetV2B1 | 80.82 | 81.11 | 80.82 | 80.84 | 69.12 |
| EfficientNetV2B2 | 81.03 | 80.74 | 81.03 | 80.88 | 68.99 |
| EfficientNetV2B3 | 81.90 | 81.07 | 81.90 | 81.33 | 70.08 |
| EfficientNetV2L | 82.54 | 82.70 | 82.54 | 82.49 | 71.84 |
| EfficientNetV2M | 82.11 | 82.13 | 82.11 | 82.06 | 70.61 |
| EfficientNetV2S | 82.97 | 83.08 | 82.97 | 82.96 | 72.15 |
| InceptionResNetV2 | 82.54 | 83.46 | 82.54 | 82.85 | 71.65 |
| InceptionV3 | 81.68 | 81.51 | 81.68 | 81.51 | 69.77 |
| MobileNet | 78.66 | 77.45 | 78.66 | 77.63 | 64.78 |
| MobileNetV2 | 82.33 | 81.46 | 82.33 | 81.64 | 70.48 |
| MobileNetV3 | 81.46 | 80.37 | 81.46 | 80.70 | 69.33 |
| NASNetLarge | 67.24 | 67.05 | 67.24 | 66.69 | 45.41 |
| NASNetMobile | 67.89 | 66.00 | 67.89 | 65.77 | 44.79 |
| ResNet101 | 79.53 | 79.24 | 79.53 | 78.85 | 66.79 |
| ResNet101V2 | 79.09 | 77.96 | 79.09 | 78.05 | 65.78 |
| ResNet152 | 74.78 | 74.59 | 74.78 | 72.73 | 58.42 |
| ResNet152V2 | 81.03 | 80.27 | 81.03 | 80.32 | 69.05 |
| ResNet50 | 80.82 | 79.58 | 80.82 | 79.54 | 68.39 |
| ResNet50V2 | 81.90 | 80.72 | 81.90 | 80.39 | 70.06 |
| VGG13 | 76.72 | 76.57 | 76.72 | 76.49 | 62.28 |
| VGG16 | 80.60 | 80.72 | 80.60 | 80.14 | 68.71 |
| VGG19 | 69.18 | 58.49 | 69.18 | 62.58 | 47.94 |
| Xception | 84.70 | 85.44 | 84.70 | 84.97 | 75.21 |

### 4.2. Results of the ViT Models

The performance of the ViT-based models was also analyzed based on various measurement metrics and results on test data were listed in Table 6.

In the DaViT series, the best result was obtained by the DaViT-Base model, with an accuracy of 83.55% and an f1-score of 83.57%. The DaViT-Tiny model performed similarly, achieving 83.41% accuracy and 83.58% f1-score. However, the DaViT-Huge model lagged behind the other models with 82.76% accuracy and 82.70% f1-score.

FastViT-T12 model performed the best among the FastViT models, achieving 84.91% accuracy and 83.85% f1-score. The FastViT-T8 model achieved a competitive result with an accuracy of 83.76% and an f1-score of 83.76%. FastViTSA-12, one of the smaller models in the FastViT series, showed a relatively low accuracy of 81.90% and an f1-score of 82.26%.

The performance analysis of the FlexiViT series models shows that the most successful model is FlexiViT-Large. With an accuracy of 82.69% and an f1-score of 82.57%, this model outperformed the other models in the series. FlexiViT-Small came in second place with a balanced performance of 82.54% accuracy and 82.50% f1-score. FlexiViT-Base performed the lowest in the series with an accuracy of 81.99% and an f1-score of 82.00%.

In the GCViT series, GCViT-Tiny model showed the highest performance, with an accuracy of 83.62% and an f1-score of 83.78%. GCViT-Small performed slightly lower with 82.54% accuracy and 82.81% f1-score.

According to the performance analysis of the GPViT series models, the most successful model was GPViT-L4. GPViT-L4 is the overall winner of the series with an accuracy of 81.90% and an f1-score of 82.19%. GPViT-L2 ranked second with 81.86% accuracy and 81.98% f1-score, performing satisfactorily in terms of both accuracy and f1-score. GPViT-L1 was the lowest-performing model in the series with 81.22% accuracy and 81.33% f1-score.

Regarding the LeViT series, the LeViT-128 model achieved the best results of the series with an accuracy of 82.70% and an f1-score of 83.20%, while the larger versions, the LeViT-256 and LeViT-384 models, underperformed with accuracies of 81.01% and 80.80%, respectively.

Among the MaxViT series models, the MaxViT-Tiny was the highest-performing model. MaxViT-Tiny achieved 86.42% accuracy and 86.53% f1-score, and MaxViT-Small achieved 84.70% accuracy and 84.95% f1-score. The larger versions of the MaxViT series, MaxViT-Base and MaxViT-Large, have lower accuracy and f1-scores of 82.33% and 81.90% accuracy, respectively.

The highest accuracy and f1-score values within the PVTV2 series were achieved by the PVTV2-B1 model with 83.84% accuracy and 83.81% f1-score. The PVTV2-B5 model also showed a remarkable performance with an accuracy of 82.97% and an f1-score of 83.40%. However, the other models of the series, especially PVTV2-B2 and PVTV2-B3, gave lower results, staying around 81% accuracy.

In the SwinTransformerV2 series, the SwinTransformerV2-Small model performed the best with an accuracy of 84.48% and an f1-score of 84.90%. SwinTransformerV2-Tiny with 84.40% accuracy and 84.35% f1-score and SwinTransformerV2-Large with 84.27% accuracy and 84.22% f1-score achieved similarly high results.

**Table 6.** Results of ViT models.

| Model | Performance measurement metrics (%) | | | | |
|---|---|---|---|---|---|
| | Accuracy | Precision | Recall | F1-score | Cohen's kappa |
| DaViT-Base | 83.55 | 83.67 | 83.55 | 83.57 | 7321 |
| DaViT-Large | 82.97 | 83.51 | 82.97 | 83.15 | 72.57 |
| DaViT-Small | 82.97 | 84.39 | 82.97 | 83.38 | 72.69 |
| DaViT-Tiny | 83.41 | 83.90 | 83.41 | 83.58 | 73.25 |
| DaViT-Huge | 82.76 | 82.78 | 82.76 | 82.70 | 71.70 |
| FastViTMA-36 | 82.11 | 82.83 | 82.11 | 82.33 | 71.22 |
| FastViTS-12 | 82.33 | 81.48 | 82.33 | 81.66 | 70.94 |
| FastViTSA-12 | 81.90 | 82.73 | 81.90 | 82.26 | 70.74 |
| FastViTSA-24 | 82.97 | 83.39 | 82.97 | 83.14 | 72.53 |
| FastViTSA3-6 | 83.62 | 83.64 | 83.62 | 83.61 | 73.21 |
| **FastViT-T12** | **84.91** | **84.34** | **84.91** | **84.62** | **74.68** |
| FastViT-T8 | 83.76 | 83.84 | 83.76 | 83.76 | 73.53 |
| FlexiViT-Base | 81.99 | 82.13 | 81.99 | 82.00 | 70.59 |
| FlexiViT-Large | 82.69 | 82.48 | 82.69 | 82.57 | 71.87 |
| FlexiViT-Small | 82.54 | 82.60 | 82.54 | 82.50 | 71.38 |
| GCViT-Base | 82.97 | 84.22 | 82.97 | 83.26 | 72.63 |
| GCViT-Small | 82.54 | 83.85 | 82.54 | 82.81 | 71.77 |
| GCViT-Tiny | 83.62 | 85.06 | 83.62 | 83.78 | 73.35 |
| GPViT-L1 | 81.22 | 81.71 | 81.22 | 81.33 | 69.25 |
| GPViT-L2 | 81.86 | 82.50 | 81.86 | 81.98 | 70.32 |
| GPViT-L3 | 80.82 | 82.84 | 80.82 | 81.27 | 69.19 |
| GPViT-L4 | 81.90 | 83.45 | 81.90 | 82.19 | 70.76 |
| LeViT-128 | 82.70 | 84.51 | 82.70 | 83.20 | 72.32 |
| LeViT-192 | 81.65 | 83.13 | 81.65 | 82.11 | 70.47 |
| LeViT-256 | 81.01 | 82.08 | 81.01 | 81.32 | 69.27 |
| LeViT-384 | 80.80 | 81.78 | 80.80 | 81.06 | 68.77 |
| MaxViT-Base | 82.33 | 83.96 | 82.33 | 82.63 | 71.91 |
| MaxViT-Large | 81.90 | 83.11 | 81.90 | 82.09 | 71.10 |
| MaxViT-Small | 84.70 | 85.71 | 84.70 | 84.95 | 75.59 |
| **MaxViT-Tiny** | **86.42** | **86.83** | **86.42** | **86.53** | **78.16** |
| PVTV2-B0 | 81.90 | 81.94 | 81.90 | 81.76 | 70.74 |
| PVTV2-B1 | 83.84 | 83.83 | 83.84 | 83.81 | 73.55 |
| PVTV2-B2 | 81.78 | 81.82 | 81.78 | 81.74 | 70.17 |
| PVTV2-B3 | 82.20 | 82.30 | 82.20 | 82.20 | 70.93 |
| PVTV2-B4 | 82.48 | 82.41 | 82.48 | 82.39 | 71.38 |
| PVTV2-B5 | 82.97 | 84.38 | 82.97 | 83.40 | 72.73 |
| SwinTransformerV2-Base | 83.69 | 83.77 | 83.69 | 83.73 | 73.50 |
| SwinTransformerV2-Large | 84.27 | 84.19 | 84.27 | 84.22 | 74.27 |
| **SwinTransformerV2-Small** | **84.48** | **85.67** | **84.48** | **84.90** | **74.99** |
| SwinTransformerV2-Tiny | 84.40 | 8434 | 84.40 | 84.35 | 74.40 |

Table 6 indicates that small-sized ViT models such as MaxViT-Tiny, SwinTransformerV2-Small, and FastViT-T12 performed the best in terms of all metrics, while some of the larger and more complex models did not. Due to the extreme complexity of large models, the risk of overfitting increases. The model requires more data as the number of parameters increases. If the model is not trained with a sufficient variety of data, it may overfit the training data and lose the ability to generalize to the test data. As a result, smaller and optimized models can perform strongly in glaucoma stage detection. In this study, the main inspiration for the rescaling and enhancing MaxViT model is the impressive performance of MaxViT models on glaucoma stage detection. Therefore, MaxViT series was considered as a backbone architecture.

### 4.3. Results of the Proposed Model

*4.3.1. Scaling The MaxViT*

The performance results of MaxViT series and the scaled version of the MaxViT are listed in Table 7. Compared to MaxViT-Base and MaxViT-Large, MaxViT-Small, the MaxViT-Tiny model obtains the best result with 86.42% accuracy and 86.53% f1-score despite having fewer parameters (31M). It shows that smaller and optimized models can learn efficiently and generalize better when they are of lower complexity.

**Table 7.** The classification performances of MaxViT models with varying scales.

| Model | Performance measurement metrics (%) | | | | | |
|---|---|---|---|---|---|---|
| | Parameter (M) | Accuracy | Precision | Recall | F1-score | Cohen's kappa |
| MaxViT-Tiny | 31 | 86.42 | 86.83 | 86.42 | 86.53 | 78.16 |
| MaxViT-Small | 69 | 84.70 | 85.71 | 84.70 | 84.95 | 75.59 |
| MaxViT-Base | 119 | 82.33 | 83.96 | 82.33 | 82.63 | 71.91 |
| MaxViT-Large | 212 | 81.90 | 83.11 | 81.90 | 82.09 | 71.10 |
| **MaxViT-Scaled** | **6.2** | **87.93** | **88.10** | **87.93** | **87.96** | **80.51** |

The scaled MaxViT model given in Table 4, MaxViT-Scaled, has the highest performance in the series with 87.93% accuracy and 87.96% f1-score, having only 6.2M parameters. There is a need to balance model size, complexity, and its interaction with the data. In our case, the dataset contains a total of 1542 images, which is relatively small. As a result, this is why small models performed better on the dataset.

As illustrated in Fig. 9, the performances of MaxViT models in classifying glaucoma stages show different sensitivities and error rates for each model. While all models perform quite strongly in the "N" class, it is noteworthy that the error rates are higher in the "A" class. In particular, the MaxViT-Tiny and MaxViT-Scaled models had higher accuracy rates in the "A" class, while the MaxViT-Small and MaxViT-Base models misclassified more cases in the "A" class. This suggests that some models have difficulty classifying advanced glaucoma cases and that the classes may be confused with each other. On the other hand, in the "E" class, all models performed consistently, showing a balanced success in detecting early stages of glaucoma. These results suggest that MaxViT models are promising for glaucoma detection, but additional optimization work is needed to improve classification accuracy in advanced cases.

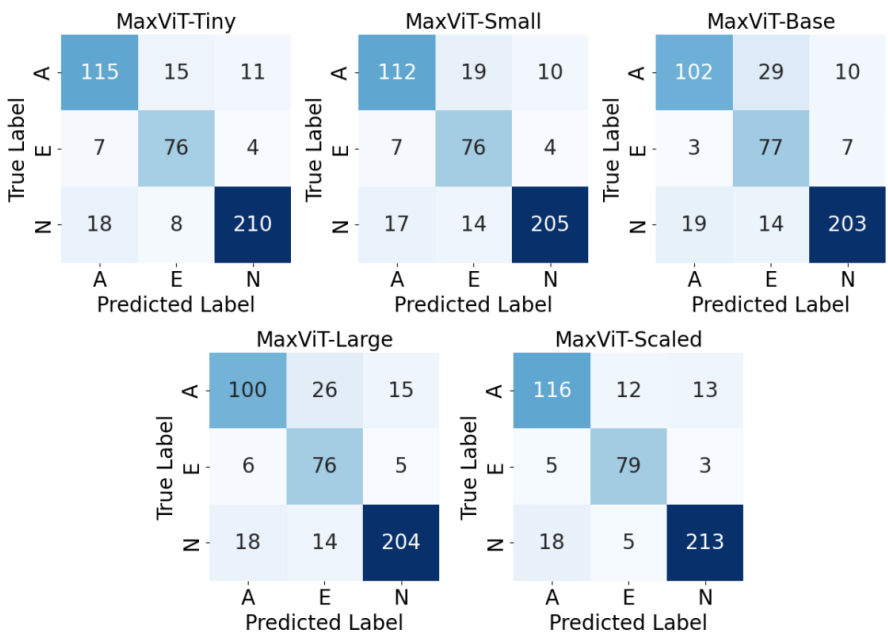

**Fig. 9.** Confusion matrices obtained with MaxViT with various scales for the glaucoma stage classification task (A: Advanced, E: Early, N: Normal).

### 4.3.2. Improved Stem

The performance results of the MaxViT-Scaled model with the stem block enhanced with different attention mechanisms are given in Table 8. The original MaxViT-Scaled model shows a superior performance with an accuracy of 87.93% and an f1-score of 87.96%. Moreover, the performance of the model is further improved by adding various attention mechanisms.

**Table 8.** Classification performance of the MaxViT-Scaled model with the stem block enhanced with different attention mechanisms.

| Model | Performance measurement metrics (%) | | | | |
|---|---|---|---|---|---|
| | Accuracy | Precision | Recall | F1-score | Cohen's kappa |
| MaxViT-Scaled | 87.93 | 88.10 | 87.93 | 87.96 | 80.51 |
| MaxViT-Scaled (ECA) | **89.01** | **89.30** | **89.01** | **89.06** | **82.32** |
| MaxViT-Scaled (CBAM) | 88.36 | 88.47 | 88.36 | 88.38 | 81.17 |
| MaxViT-Scaled (SE) | 88.15 | 88.66 | 88.15 | 88.27 | 81.00 |

The MaxViT-Scaled model with ECA achieved the highest accuracy of 89.01% and f1-score of 89.06%. Since ECA provides channel-based attention, it allows the model to better select particularly important features and neglect unimportant information. The MaxViT-Scaled model equipped with CBAM achieved an accuracy of 88.36% and f1-score of 88.38%, lower than ECA but higher than the scaled model. The MaxViT-Scaled model with the addition of the SE block performs similarly to CBAM, with an accuracy of 88.15% and an f1-score of 88.27%. It is obvious that the MaxViT-Scaled model with ECA obtained superior performance compared to other attention modules. For visual understanding, Fig. 10 shows the structure of the stem block improved with ECA.

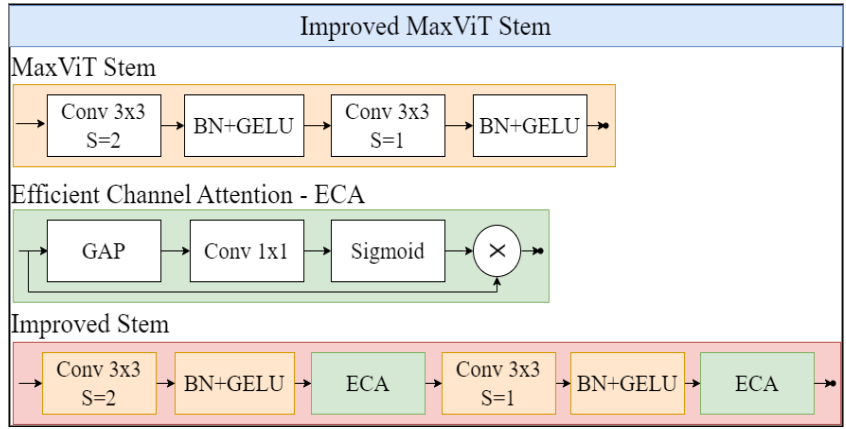

**Fig. 10.** The structure of the improved stem block.

*4.3.2. Improved MaxViT Block*

The experimental results obtained by replacing MBConv in the MaxViT block with state-of-the-art convolution modules are given in Table 9. The ConvNeXt module achieved better results than the MaxViT-Scaled model with an accuracy of 88.15% and an f1-score of 88.16%. The advanced structure of the ConvNeXt block provided a small but significant improvement to the model. The ConvNeXtV2 module is the highest performing model with an accuracy of 89.87% and f1-score of 89.93%. ConvNeXtV2 block significantly improved the performance of the MaxViT-Scaled model and extracted features from the data more effectively. The optimized structure of the

ConvNeXtV2 block compared to the previous generation ConvNeXt structure and its GRN normalizes features along the feature map, allowing the network to learn more meaningful and important features. The MaxViT-Scaled-InceptionNeXt model's performance is close to but lower than ConvNeXt with 88.77% accuracy and 88.85% f1-score. The InceptionNeXt block gave the model a broader perspective, allowing it to extract features at different scales, but it was not as successful as ConvNeXtV2, although it achieved better results than MaxViT-Scaled. The results show that MaxViT-Scaled model with ConvNeXtV2 obtained better performance compared to other convolutional blocks. For visual understanding, Fig. 11 depicts the structure of the MaxViT block improved with ConvNeXtV2.

**Table 9.** Classification performance of the MaxViT-Scaled model replacing MBConv in the MaxViT block with state-of-the-art convolution modules.

| Model | Performance measurement metrics (%) | | | | |
|---|---|---|---|---|---|
| | Accuracy | Precision | Recall | F1-score | Cohen's kappa |
| MaxViT-Scaled-MBConv | 87.93 | 88.10 | 87.93 | 87.96 | 80.51 |
| MaxViT-Scaled-ConvNeXt | 88.15 | 88.27 | 88.15 | 88.16 | 80.77 |
| **MaxViT-Scaled-ConvNeXtV2** | **89.87** | **90.23** | **89.87** | **89.93** | **83.65** |
| MaxViT-Scaled-InceptionNeXt | 88.77 | 89.06 | 88.77 | 88.85 | 81.75 |

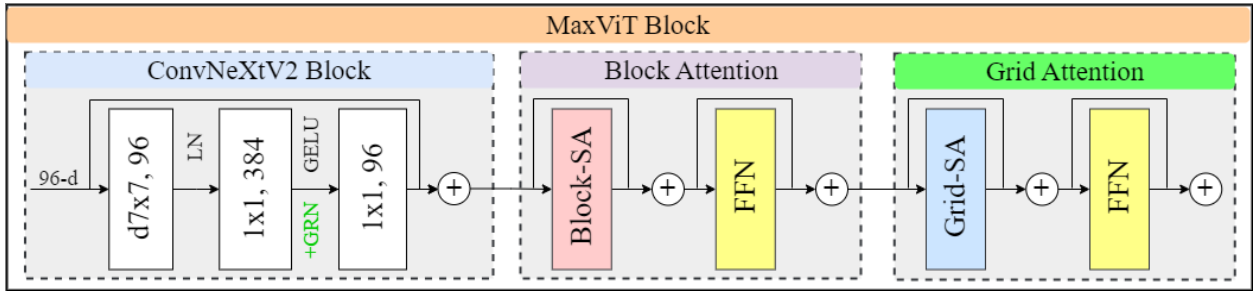

**Fig. 11.** The structure of the improved MaxViT block.

### 4.3.3. Improved Stem and MaxViT Block

At this stage of the study, three different experiments were conducted. In the first experiment, scaling the MaxViT model reduced its size (block and channel) and improved its performance. In the second experiment, the stem block was enhanced with ECA, which resulted in a more robust feature extraction and improved performance. In the last experiment, the performance was improved by adding different convolutional blocks to the MaxViT block. Considering the results of the previous experiments, the power of the ECA attention module and ConvNeXtV2 convolutional block integrated to the scaled MaxViT model as a best combination. We also experimented ECA with other convolutional blocks to prove the consistency of the proposed approach. The experimental results show the effects of combinations on the model and the results are presented in detail in Table 10.

**Table 10.** Classification performance of the MaxViT-Scaled model replacing MBConv in the MaxViT block with convolution modules and the stem block enhanced with different attention mechanisms.

| Model | Performance measurement metrics (%) | | | | |
|---|---|---|---|---|---|
| | Accuracy | Precision | Recall | F1-score | Cohen's kappa |
| MaxViT-Scaled-ECA-MBConv | 89.01 | 89.30 | 89.01 | 89.06 | 82.32 |
| MaxViT-Scaled-ECA-ConvNeXt | 90.73 | 90.85 | 90.73 | 90.78 | 84.98 |
| **MaxViT-Scaled-ECA-ConvNeXtV2** | **92.03** | **92.33** | **92.03** | **92.13** | **87.12** |
| MaxViT-Scaled-ECA-InceptionNeXt | 91.16 | 91.58 | 91.16 | 91.30 | 85.76 |

Figure 12 details the structure of the proposed MaxViT model (MaxViT-Scaled-ECA-ConvNeXtV2), named MaxGlaViT. The model starts with a stem block enhanced with ECA blocks, followed by MaxViT blocks in four stages. The structure is designed to enhance feature extraction and improve the performance of the model. The MaxViT block includes ConvNeXtV2, block and grid attention modules, and the combination of these components increases the learning capacity of the model. In the final stage, the output is generated with a pooling layer and a fully connected layer.

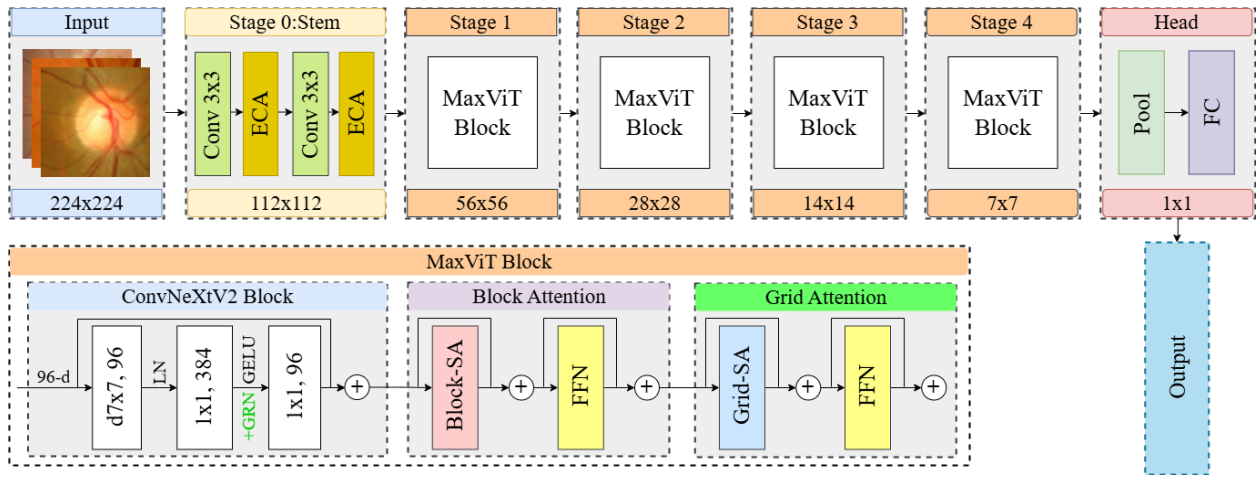

**Fig. 12.** The structure of the proposed MaxGlaViT.

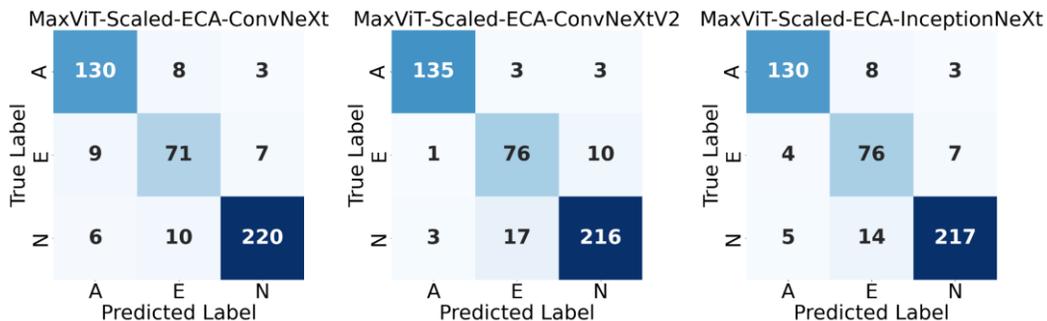

**Fig. 13.** Confusion matrices of the models (A: Advanced, E: Early, N: Normal).

Overall, all three models perform well with high correct classifications, especially for class "N", where each model achieves over 215 correct predictions shown in Fig. 13. MaxViT-Scaled-ECA-ConvNeXtV2 has the highest accuracy for class "A", with 135 correct classifications, indicating its strength in recognizing advanced glaucoma stages. For class "E", MaxViT-Scaled-ECA-ConvNeXtV2 and MaxViT-Scaled-ECA-InceptionNeXtV2 have the fewest misclassifications, suggesting it is more effective at distinguishing class E from the others. These insights suggest that, while all models are competent, MaxViT-Scaled-ECA-ConvNeXtV2 is preferable for glaucoma stage detection.

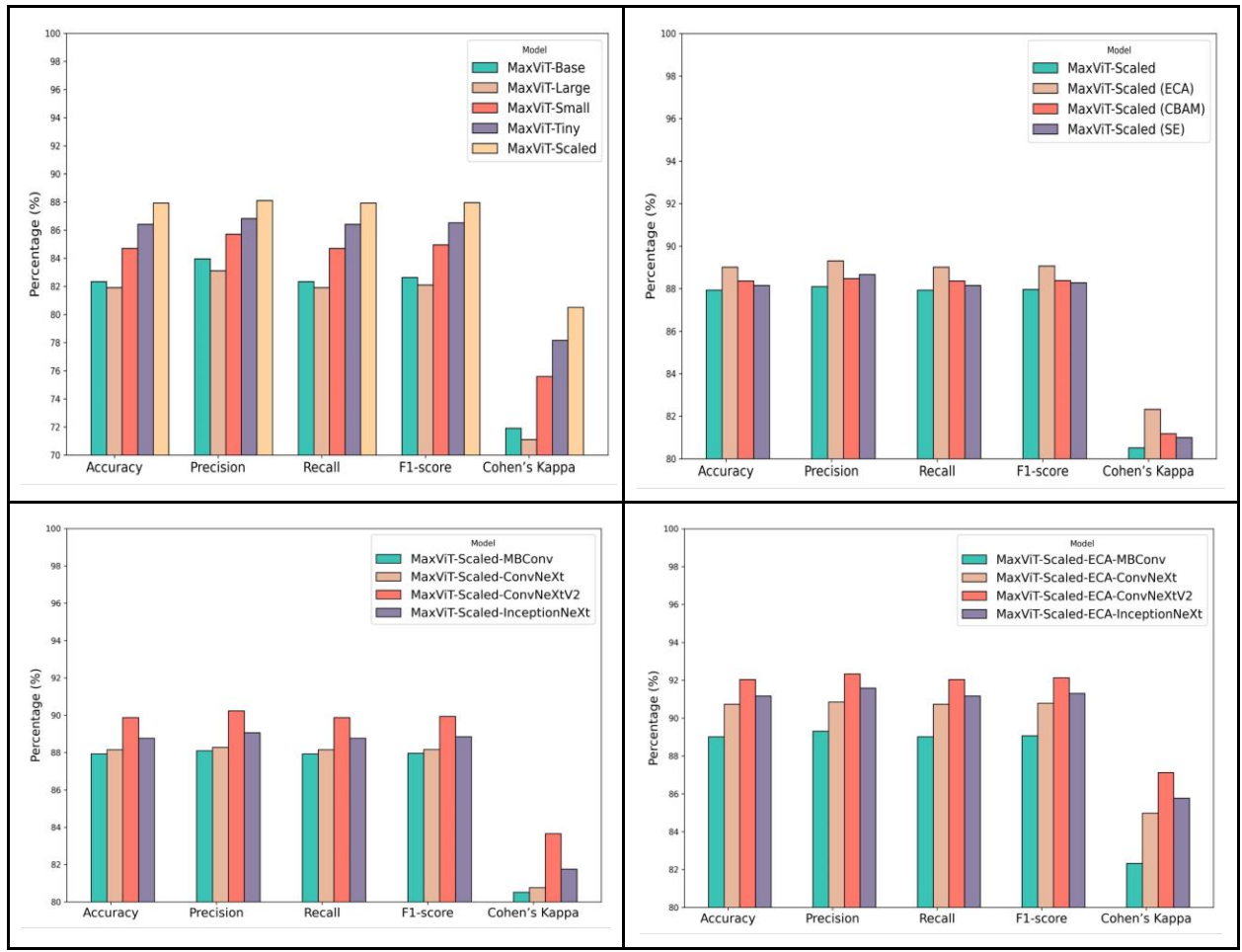

**Fig. 14.** Bar graph for results of the modified MaxViT deep models with different variations.

As visually depicted in Fig. 14, overall, MaxViT-Scaled-ECA-ConvNeXtV2 consistently outperforms the other variations across all metrics, establishing itself as the most effective model in this comparison. The balanced architecture of the model likely contributes to its ability to generalize well across diverse tasks, ensuring robust performance. MaxGlaViT is not only as the best among the tested models but also as a promising candidate for broader applications in real-world scenarios.

### 4.4. Comparison of the proposed model with other literature studies

The performance of the MaxGlaViT model is evaluated by comparing it with other studies in the literature using the same dataset. Based on Table 11, the proposed MaxGlaViT model achieved significant success in the field of fundus image-based glaucoma detection, reaching 92.03% accuracy, 92.33% precision, 92.03% recall, and 92.13% f1-score. Compared to FJA-Net, which has the highest performance in the literature, MaxGlaViT increased its accuracy from 87.06% to 92.03%, an increase of 5.71%. The precision improved by 6.11%, from 87.01% to 92.33%, and the recall improved by 5.71%, from 87.06% to 92.03%. Also, the f1-score increased from 86.90% to 92.13%, an increase of 6.02%. For healthcare, even a 1% increase in accuracy is significant enough to make a critical difference in accurate diagnosis and treatment.

**Table 11.** The comparison of MaxGlaViT with other literature studies that use the same dataset.

| Model | Performance measurement metrics (%) | | | |
| --- | --- | --- | --- | --- |
| | **Accuracy** | **Precision** | **Recall** | **F1-score** |
| InceptionV3 [9] | 84.50 | - | - | - |
| AES-Net [14] | 86.20 | 85.32 | 85.77 | 85.46 |
| FJA-Net [15] | 87.06 | 87.01 | 87.06 | 86.90 |
| CA-Net [16] | 85.34 | 85.15 | 85.34 | 84.92 |
| GS-Net [17] | 84.91 | - | - | 84.55 |
| **Proposed method (MaxGlaViT)** | **92.03** | **92.33** | **92.03** | **92.13** |

The results indicate that MaxGlaViT provides a significant performance advantage compared with other studies in the literature. In FJA-Net, AES-Net, GS-Net, and CA-Net, a CNN model is used as a backbone and an attention mechanism is added to the last layer for classification. However, one study uses only InceptionV3 and transfer learning. CNNs struggle to capture long-range contextual information, as they focus primarily on local feature extraction, which may limit the ability to understand global context, leading to reduced robustness in image classification. However, MaxViT has an innovative architecture that can effectively learn both local and global contextual information. With its grid and block attention mechanisms, MaxViT is capable of learning long-range dependencies where CNNs are limited. In addition to the inherited features in MaxViT, MaxGlaViT model has learned many local and global features thanks to the improvements in the stem and MaxViT block, and outperforms recent studies in the literature.

### 4.5. Discussions

The paper introduces an enhanced MaxViT-based model, MaxGlaViT, for the classification of glaucoma stages from fundus images (Fig. 7). The model is designed by rescaling the MaxViT architecture, which provides an effective trade-off between performance and model size. The Scaled MaxViT has 6.2M parameters, 80% fewer than the MaxViT-Tiny model. MaxViT-Scaled achieved 87.93% accuracy, 88.10% precision, 87.93% recall, 87.96% f1-score and 80.51% Cohen's kappa (Table 7). The results indicate that fine-tuning model structure complexity can improve generalizability, especially when working with limited data. Further improvements were achieved with attention mechanisms integrated into the stem block. The addition of ECA to the stem block achieved 89.01% accuracy, 89.30% precision, 89.01% recall, 89.06% f1-score, and 82.32% Cohen's Kappa (Table 8). This enhancement enabled channel-based features to be emphasized in early-stage feature extraction. Replacing the MBConv block in the MaxViT block with the ConvNeXtV2 module resulted in 89.87% accuracy, 90.23% precision, 89.87% recall, 89.93% f1-score, and 83.65% Cohen's kappa (Table 9). Finally, in the scaled model, using ConvNeXtV2 in the ECA and MaxViT blocks in the stem block and ECA in the MaxViT block, an accuracy of 92.03%, precision of 92.33%, recall of 92.03%, f1-score of 92.13%, and Cohen's Kappa of 87.12% were obtained (Table 10). Comparisons with existing literature demonstrate that the MaxGlaViT model is superior in terms of classification performance (Table 11). Finally, proposed MaxGlaViT (Fig. 12) demonstrates notable performance, achieving 92.03% accuracy, 92.33% precision, 92.03% recall, 92.13% f1-score, and 87.12% Cohen's kappa score compared to existing over 80 deep models and models in mentioned literature studies.

## 5. Conclusion

Glaucoma is a chronic eye disease and it leads to irreversible vision loss if diagnosed at an early stage. This paper presents a CAD system to assist ophthalmologists in the diagnostic process of glaucoma stages. Experiments were performed on three main DL models, namely CNNs, ViTs and MaxGlaViT. Among 40 CNN models, EfficientB6 was the most successful model with an accuracy of 84.91%. On the other hand, among the ViT models, MaxViT-Tiny achieved the highest performance with an accuracy rate of 86.42%. We then scaled the number of blocks and channels of the MaxViT-Tiny, resulting in a lightweight model with 6.2M parameters and an accuracy rate of 87.93%. After adding ECA to the stem block, the accuracy rate increased to 89.01%. Another improvement was made by replacing the MBConv structure in the MaxViT block with ConvNeXtV2 and an accuracy of 89.87% was obtained. In the last stage, the MaxGlaViT model was obtained by using ECA and ConvNeXtV2 block on the stem and MaxViT block, respectively, and the accuracy was increased to 92.03%. Experimental results prove that the proposed lightweight MaxGlaViT model is among the most advanced models in this field by showing superior performance in glaucoma diagnosis. In future work, potential improvements and mechanisms that can be applied to the block, grid attention and head parts of the MaxViT model will be investigated and experiments will be conducted.

**Declaration of competing interest**

The authors declare that they have no known competing financial interests or personal relationships that could have appeared to influence the work reported in this paper.

**CRediT authorship contribution statement**

**Mustafa Yurdakul, Kübra Uyar:** Conceptualization, Methodology, Review and editing, Software, Validation, Visualization, Writing-original draft, **Şakir Taşdemir:** Conceptualization, Methodology, Supervision.

**Code Availability** The source code used to obtain experimental results is publicly available at https://github.com/ymyurdakul/MaxGlaViT.

**Data Availability** The data that support the findings of this study are openly available at https://dataverse.harvard.edu/citation?persistentId=doi:10.7910/DVN/1YRRAC.